\PassOptionsToPackage{pdftex,dvipsnames}{xcolor}
\documentclass[letterpaper, 10 pt, conference]{ieeeconf}  

\IEEEoverridecommandlockouts                             
\usepackage{graphics}
\usepackage{epsfig} 
\usepackage{mathptmx} 
\usepackage{times} 
\usepackage{amsmath} 
\usepackage{amssymb}  
\usepackage{subcaption}
\usepackage{graphicx}
\usepackage{float}
\usepackage{bm}
\usepackage{caption}
\usepackage{mathtools}
\usepackage{stfloats}
\usepackage{mathrsfs}
\usepackage{multirow}
\usepackage{makecell}
\usepackage{vcell}
\usepackage{booktabs}
\usepackage{diagbox}
\usepackage{csquotes}
\usepackage{cite}
\usepackage{tikz}
\usepackage{lipsum}
\usepackage{schemata}
\usepackage{verbatim}
\usepackage{pifont}
\newcommand{\cmark}{\ding{51}}%
\newcommand{\xmark}{\ding{55}}%

\usepackage[hyphens]{url}
\usepackage[hidelinks]{hyperref}
\hypersetup{breaklinks=true}
\usepackage{cite}

\usepackage{algorithm}
\usepackage[noend]{algpseudocode}

\algrenewcommand\algorithmicforall{\textbf{for each}}
\algrenewcommand\algorithmicindent{.8em}

\algdef{SE}[DOWHILE]{Do}{doWhile}{\algorithmicdo}[1]{\algorithmicwhile\ #1}%
\algnewcommand\algorithmicforeach{\textbf{for each}}
\algdef{S}[FOR]{ForEach}[1]{\algorithmicforeach\ #1\ \algorithmicdo}

\usetikzlibrary{decorations.pathreplacing,calc}

\usepackage{hyperref}
\hypersetup{
colorlinks=true,
linkcolor=blue,
filecolor=magenta,
urlcolor=blue,
}
\usepackage[font=footnotesize]{caption}

\title{\LARGE \bf
DynaCon: Dynamic Robot Planner with Contextual Awareness via LLMs
}
\author{Gyeongmin Kim$^{1}$$\dagger$, Taehyeon Kim$^{2}$$\dagger$, Shyam Sundar Kannan$^{2}$, Vishnunandan L. N. Venkatesh$^{2}$,\\ Donghan Kim$^{3}$ and Byung-Cheol Min$^{2}$
\thanks{$^{1}$The author is with Helper Lab, Department of Intelligent Robotics, SungKyunKwan University, Suwon 16419, Republic of Korea \tt\small{hn04008@g.skku.edu}}%
\thanks{$^{2}$The authors are with SMART Lab, Department of Computer and Information Technology, Purdue University, West Lafayette, IN 47907, USA \tt\small{\{kim4435, kannan9, lvenkate, minb\}@purdue.edu}}%
\thanks{$^{3}$The author is with HRI Lab, Department of Electrical Engineering, Kyung Hee University, Yongin 17104, Republic of Korea \tt\small{donghani@khu.ac.kr}}%
\thanks{$\dagger$ Equal contribution}
}

\begin{document}
\maketitle
\thispagestyle{empty}
\pagestyle{empty}

\begin{abstract}

Mobile robots often rely on pre-existing maps for effective path planning and navigation. However, when these maps are unavailable, particularly in unfamiliar environments, a different approach become essential. This paper introduces DynaCon, a novel system designed to provide mobile robots with contextual awareness and dynamic adaptability during navigation, eliminating the reliance of traditional maps. DynaCon integrates real-time feedback with an object server, prompt engineering, and navigation modules. By harnessing the capabilities of Large Language Models (LLMs), DynaCon not only understands patterns within given numeric series but also excels at categorizing objects into matched spaces. This facilitates dynamic path planner imbued with contextual awareness. We validated the effectiveness of DynaCon through an experiment where a robot successfully navigated to its goal using reasoning.
Source code and experiment videos for this work can be found at: \url{https://sites.google.com/view/dynacon}.

\end{abstract}


\section{Introduction}
\label{sec:intro}
Humans commonly refer to maps, to gauge the distance from their current location and to their intended destination to find a path. Similarly, a mobile robot acquires data about a given environment beforehand, determines its location, and navigates to its endpoint \cite{pathplanalgo}.
However, challenges arise when the robot's performance is suboptimal, or when the map is not unavailable or contains noise, making standard path planning difficult \cite{drlnav}. In such situations, it is imperative for the robot to infer unknown information using its memory and predict the desired goal's position.
This inference is often called \textbf{Contextual Awareness} in navigation.

Recently, the Large Language Model (LLM) has garnered attention due to its proficiency in reading and generalizing input sentence contexts.
Inspired by the rising trend of LLM applications across various studies, we tackle harnessing LLM for context-aware robot navigation. Our objective is to integrate LLM into a mobile robot, empowering it to understand the context of nearby objects by processing information in sentence form.
Nevertheless, it is essential to prioritize real-time object detection and responses to environmental changes for successful navigation without collision~\cite{realtimepathplan}\cite{mobilenavsensor}.
Therefore, our ultimate aim is to equip the mobile robot with the ability to comprehend its surroundings, thereby enabling efficient navigation towards its destination, even when deprived of pre-existing world information.

\begin{figure}[t]
\centering
\includegraphics[width=0.87\columnwidth]{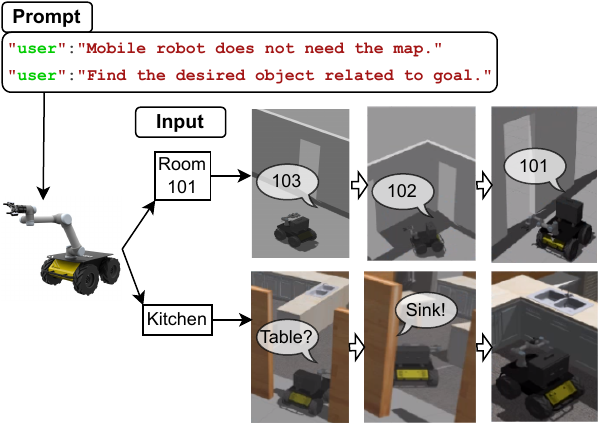}
\vspace{-5pt}
\caption{Given input sentences, DynaCon can estimate goals through either \textit{pattern-based reasoning} (top: identifying patterns within numerical sequences) or \textit{categorical reasoning} (bottom: categorizing objects into specific rooms). In both scenarios, DynaCon can execute navigation with periodic updates in detection, even in the absence of a map.}
\vspace{-17pt}
\label{fig1}
\end{figure}

In this work, we introduce DynaCon, Dynamic path planner with Contextual awareness via LLM (Fig.~\ref{fig1}).
This path planner continually updates the list of surrounding objects whenever changes occur in their presence, thus significantly improving the accuracy of real-time planning for the mobile robots.
More importantly, it empowers the mobile robot to engage in contextual navigation, enabling it to reach its destination in an unknown environment.
To enhance the versatility of DynaCon, we employ two distinct approaches which are pattern recognition or \textit {pattern-based reasoning} and classification or \textit {categorical reasoning} for contextual estimation.
The contributions of this paper are as follows:
\begin{itemize}

 \item We propose DynaCon, an innovative framework that integrates rapid engineering, real-time feedback, and navigation operations. DynaCon empowers effective robot navigation, even in challenging unknown and dynamic environments.

 \item We introduce a prompt engineering structure, categorized into Role, Main Task, and Instruction components. This structure enhances the learning capabilities of LLM, making it adept at handling complex information.

 \item We craft our prompts for LLM to enable reasoning in both \textit{pattern-based} and \textit{categorical} manners. This approach enhances LLM's ability for generalized robot navigation, bolstering its adaptability and effectiveness across diverse scenarios.
\end{itemize}


\section{Related Works}
\label{sec:rel_work}

\begin{figure*}[t]
\includegraphics[width=\linewidth]{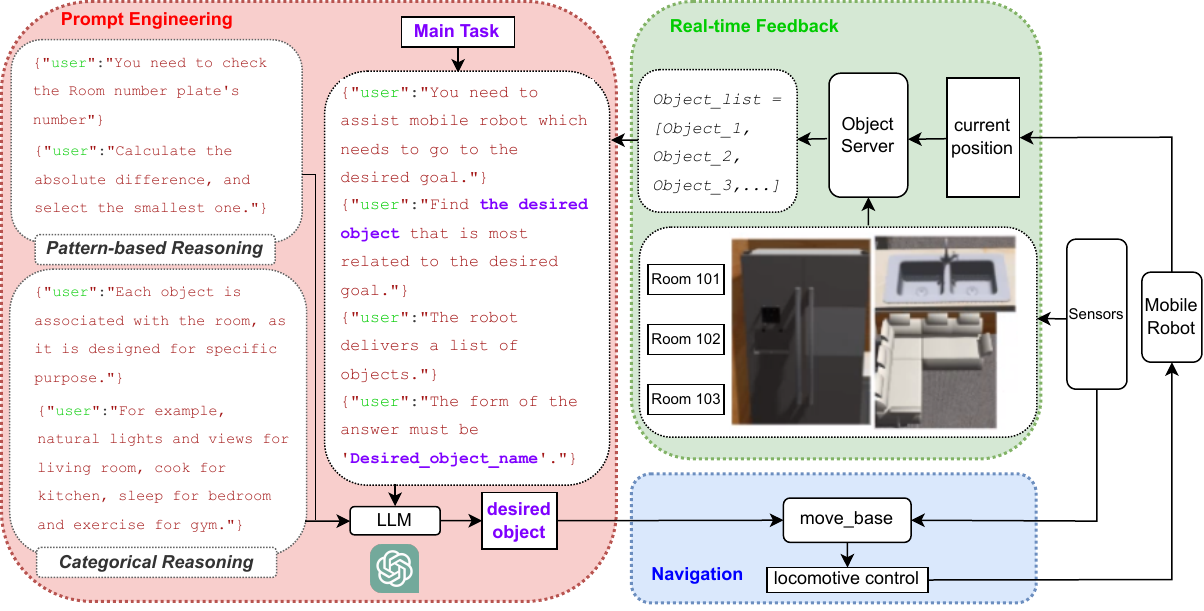}
\vspace{-15pt}
\caption{An overview of DynaCon's structure: Within the subsections of real-time feedback, prompt engineering, and navigation task, DynaCon processes task inputs, recognizes objects from the environment in real-time, and performs navigation towards the desired object by reasoning based on designed prompts. In the \textcolor{green}{real-time feedback} section, the Object Server periodically retrieves information about nearby objects and the current position of the robot to update the object list. In the \textcolor{red}{prompt engineering} phase, a uniquely structured prompt is sent to the Large Language Model (LLM) to output the desired object, serving as the main task of navigation. Finally, the ROS \textit{move\textunderscore base} applied for \textcolor{blue}{navigation}.}
\vspace{-7pt}
\label{fig2}
\end{figure*}

\noindent\textbf{Context-Aware Navigation in Robotics.} As the interest in Human-Robot Interaction (HRI) has grown, facilitated by the decreasing distance between humans and robots \cite{commuhri}, the ability of the robot to recognize context has become crucial. Some researchers, such as \cite{ped1}\cite{ped2}, have focused on the context of pedestrians, demonstrating the robot's capability to navigate non-threateningly in dynamic obstacle-rich indoor environments. Other studies, like \cite{inter1}\cite{inter2}, emphasize interactive contexts involving direct interactions between robots and humans. They explore how to translate natural language from humans into the robot's programming language spontaneously. Meanwhile, researchers like \cite{local1}\cite{local2} view context in terms of local context. They aim to infer unknown place labels and achieve more efficient navigation by integrating simultaneous localization and mapping (SLAM) rather than relying solely on manual goal judgment.

\noindent\textbf{Robot Planning via LLM.} In the pursuit of robot automation, various methods have been introduced to address the challenge of planning via LLM~\cite{saycan, microsoft, navcon, VLMaps, progprompt, llmdp, capeam}. SayCan~\cite{saycan} demonstrates the feasibility of completing tasks using LLMs.
With value functions of pre-trained data and semantic knowledge sources from LLMs, it can perform tasks described in long sentences. The work in~\cite{microsoft} showcases the application of OpenAI's ChatGPT, a well-known chatbot based on the GPT model, to a range of robotic tasks.
Leveraging well-designed prompts and APIs, this system can control drones and manipulators, applying spatio-temporal reasoning. 

NavCon~\cite{navcon}, which serves as an intermediary layer between LLM and navigation, generates Python code for navigation, integrating  natural language, visual information, and maps. 
VLMaps~\cite{VLMaps} combines visual-language models with 3D map representations to generate waypoints using natural language.
Thanks to the code-generation capabilities of LLMs, it can recognize spatial objectives even in ambiguous statements. However, its dependence on visual input is a noted limitation.

PROGPROMPT~\cite{progprompt} reveals that the text generalization ability of LLM can be harnessed for task planning based on well-structured prompts.
It also demonstrates how embedding comments within prompts improves the LLM's performance. Both LLM-DP~\cite{llmdp} and CAPEAM~\cite{capeam} share considerations similar to ours, aiming to conduct context-aware planning dynamically with LLM.
However, DynaCon uniquely manges the context of patterns or spatial categories derived from unseen maps, as well as from command sentences via structured prompt engineering.


\section{Methodology}
\label{sec:methodology}
DynaCon consists of three parts: real-time feedback, prompt engineering on LLM, and navigation with Robot Operating System (ROS), as depicted in Fig.~\ref{fig2}.

\begin{figure}[t]
\includegraphics[width=\linewidth]{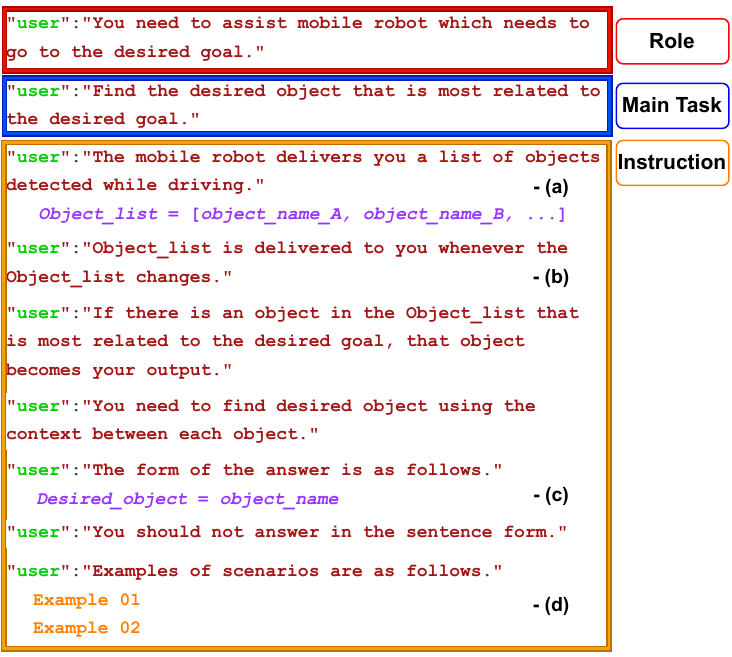}
\caption{An example of prompt engineering with LLM. The \textcolor{red}{red} box represents the role component, the \textcolor{blue}{blue} represents the main task, and \textcolor{orange}{orange} box signifies the instruction for limiting the reasoning boundary. Within the \textcolor{orange}{orange} box, (a)-(d) describe specific constraints: (a) illustrates the format of the received object list, (b) denotes the real-time situation, (c) outlines the prerequisites and desired output format for objects, and (d) provides scenario examples for the reasoning process.}
\vspace{-10pt}
\label{fig3}
\end{figure}

\subsection{Real-time Feedback on Object Server}
Initially, the object server processes two inputs: the current location of the mobile robot and the object information from the environment. For object detection, it is assumed that the sensor can identify objects within a specific range to enable DynaCon's role in navigation.
The object server then outputs an \textit {object list}, which is a Python list populated with the names of detected objects.
It is important to note dynamic nature of the object server. As the reference point from which the sensor measures continually shifts, the elements (name) within the object list change over time.
Consequently, if the server identifies new objects or some previously detected objects are no longer present, it will periodically update the object list and feed it as input to the LLM.

\subsection{Prompt Engineering on LLM}
Prompts fed to the LLM can be categorized into two types based on their role: reasoning and navigation.
Reasoning prompts, provided prior to the actual navigation, train the LLM on the scenarios it may encounter. THey furnish examples to help the LLM deduce patterns or appropriate categories.
Importantly, a single shot of such a prompt aids in preserving the reasoning processing throughout the entire navigation task. 
To optimize performance, we employ prompt engineering to design a precise question format, which encompasses both input values. 
Fig.~\ref{fig3} displays an example reasoning prompt fashioned using our recommended design, which includes \textbf{Role}, \textbf{Main Task}, and \textbf{Instruction}. This design is aimed at enhancing the LLM's capabilities.

\noindent\textbf{Role:} This is designed to prescribe a specific function to the LLM. Through this element, we determine which domain of knowledge the LLM should tap into by framing a particular context of perspective. For example, by positioning the LLM as an AI assistant for a mobile robot aiming to reach a destination, the LLM conceptualizes its role based on the provided context. 

\noindent\textbf{Main Task:} This is to define the primary objective for the LLM.
Since the response is shaped by this directive, it is imperative that the instruction is unambiguous and succinct, utilizing domain-specific terms when necessary. In this context, the main objective is identifying the object most pertinent to the desired destination. 

\noindent\textbf{Instruction:} This is to delineate the scope of LLM's action.
Given LLM's reliance on extensive datasets as its information source, this instruction imposes limitations on its actions. Consequently, as a constraint, we present a desired format to structure the answer or reduce the size of the dataset by describing the current context. 
Additionally, we provide LLM with an example scenario to guide its reasoning in a specific direction when addressing incoming queries.
This scenario contains rules related to two distinct types of reasoning, \textit{pattern-based reasoning} and \textit{categorical reasoning}.
In \textit{pattern-based reasoning}, DynaCon discerns patterns within a series of input numbers. 
For instance, if presented with a decreasing pattern, it can predict that the position of room number 101 will be farther than 102 when the robot is currently located at room 103.
On the other hand, employing \textit{categorical reasoning}, DynaCon categorizes each object based on its specific role within an environment.
For example, it can estimate that it needs to locate and move to the sink to reach the kitchen or to the sofa to reach the living room.
Following the application of a well-structured prompt to guide LLM's  reasoning, we assign the main task of navigation via the prompt ,which is a simple sentence of \texttt{Go to the place}.
LLM, trained with semantic data, receives this main task and provide the the desired object based on the updated object list.

\subsection{Navigation Task with ROS}
The terminal point for ROS navigation is determined based on the name and distance of the desired object from the mobile platform. We leverage the pre-existing \textit{move\textunderscore base} package within the navigation stack and utilize \textit{Navfn}, which is based on Dijkstra's algorithm  \cite{navfn}, as the global path planner. For local path planning, we implement dynamic window approach (DWA) \cite{dwa}. The loop is closed by sending control values to the mobile robot, ensuring effective and dynamic navigation towards the target.


\section{Experiments}
\label{sec:experiments}

\subsection{Ablation Study for Prompt Engineering}
\begin{figure}[t]
\includegraphics[width=\linewidth]{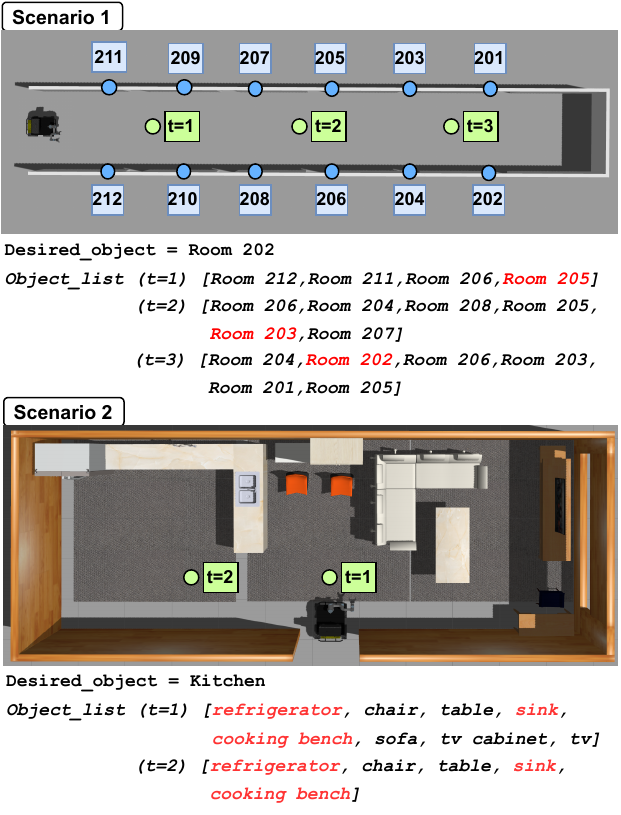}
\caption{Two scenarios designed for the ablation study on prompt engineering. Scenario 1 is based on patterns, and Scenario 2 is based on categories. The text highlighted in \textcolor{red}{Red} represents the object that the prompt should select based on the reasoning rules.}
\vspace{-13pt}
\label{fig4}
\end{figure}

Before conducting experiments, we evaluate the effectiveness of our prompt design presented in Section \ref{sec:methodology}.
The unstructured prompt, which contains the information necessary for context-aware navigation but lacks the refinement of prompt engineering, is as follows.
\begin{itemize}
	\item \textsl{Find the desired object using context-aware navigation}.
        \item \textsl{The desired object is within the object list most relevant to the desired goal}.
        \item \textsl{The object list is provided whenever it undergoes changes}.
        \item \textsl{If a room number is obtained from the room number plate, calculate the absolute difference between the desired goal and each room number plate. Select the one with the smallest absolute difference as the desired object}.
        \item \textsl{If a specific room is specified, identify the object most suitable for categorization within the given room}.
        \item \textsl{Going forward, I will provide you with my desired goal and an updated object list. Show me your process for reaching the desired goal}.
\end{itemize}
Following this, we present two scenarios,based on patterns and categories, which will be utilized to evaluate the performance of DynaCon in the experiments.
Fig.~\ref{fig4} represents detailed information on these two scenarios.
Based on the rules outlined in the essential information, we can infer that the robot should select Room 205, 203, and 202 at different times (i.e., t$=1$, t$=2$, t$=3$) to fulfill the pattern in Scenario 1. In Scenario 2, the robot should target the refrigerator, sink, or cooking bench to reach the kitchen. We use the same LLM model, GPT-3.5, for both the engineered and unstructured prompts.

\subsection{Experiments}
\begin{figure*}[t]
\includegraphics[width=\linewidth]{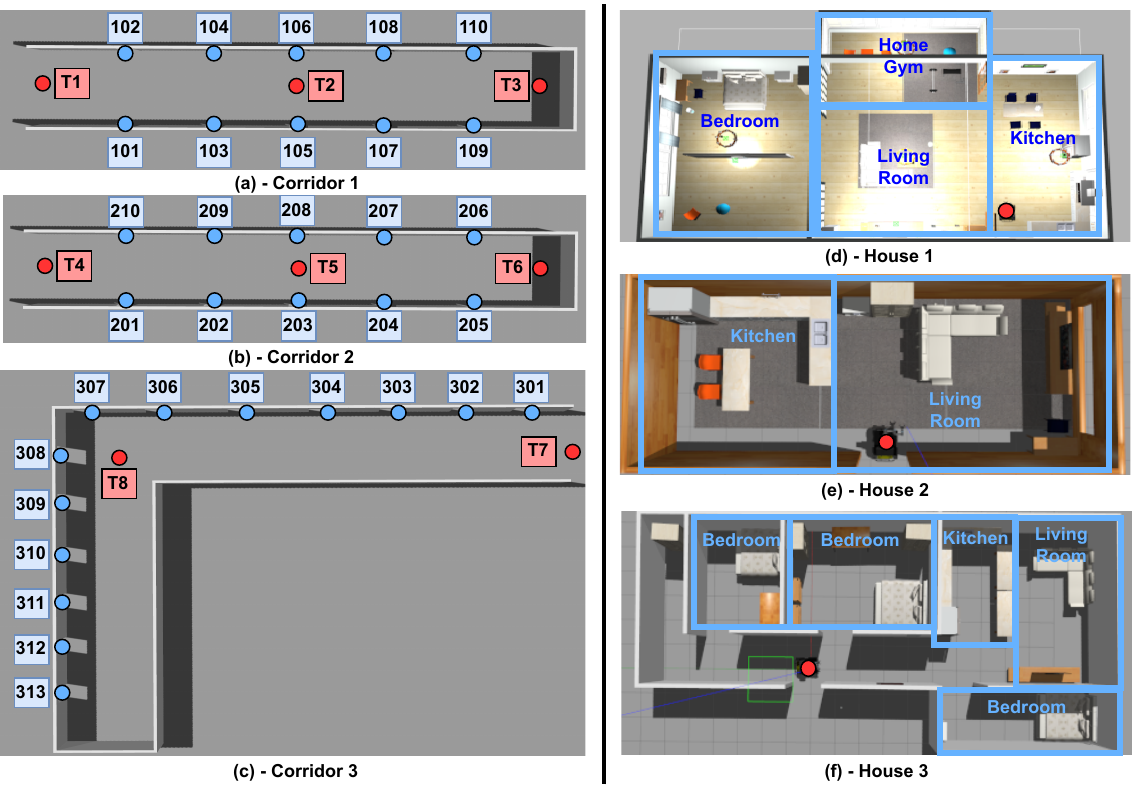}
\vspace{-10pt}
\caption{Simulated environments for evaluating different forms of reasoning. The corridors (a)-(c) represent scenarios for \textit{pattern-based reasoning}, while the houses (d)-(f) depict settings for \textit{categorical reasoning}. The \textcolor{red}{Red} points indicate the initial positions of the mobile robot manipulator in each setting.}
\vspace{-5pt}
\label{fig5}
\end{figure*}

We aim to assess whether the mobile manipulator equipped with DynaCon can effectively perform real-time feedback and reasoning via LLM during navigation. 
We simulate two types of reasoning tasks, namely \textit {pattern recognition} and \textit{categorization}, using GPT-3.5 model.
Fig.~\ref{fig5} represents an overview of the entire set of simulation experiments for each task.

\subsubsection{Pattern-based Reasoning}
We conduct a total 8 experiments across three virtual worlds.
These environments include two simple linear (Corridor 1 and 2) and one L-shaped hallway with a single perpendicular corner (Corridor 3). 
In Corridor 1 and 2, we position three different starting points at each end and the middle of the straight aisles.
In Corridor 3, we designate two starting points at one end and the center of the corner.
Each of these environments contains 10 doors in Corridor 1 and 2, and 13 doors in Corridor 3, each with distinct room numbers. 
Specifically, for Corridor 1 and 2, we present two specific room number patterns: monotonically increasing and decreasing, and counter-clockwise, respectively.
Corridor 3 exhibits either an increasing or decreasing order on one side of aisle, depending on its initial point.
In this case, the robot receives input regarding a room number it does not know in advance and navigates to its goal based solely on inference.

\subsubsection{Categorical Reasoning}
In the task of categorizing objects, we prepare 3 different house maps.
These environments are randomly chosen, representing various residential environments. 
Here, we assess whether DynaCon can classify objects into each room based on their function and navigate to the target room when prompted.
It is essential to note that each connection between rooms and objects is not repeatably trained, except for the initial training with scenarios.
All 8 experiments within the 3 houses start from the same initial point, the front door.
We set the goals in various room areas, including the kitchen, living room, home gym, and bedroom.

\subsubsection{Evaluation Metrics}
In both experiments, we evaluate whether the robot successfully reaches its desired goal.
If it fails to identify the desired object or ends up in an unrelated location, the task is considered a failure
We calculate the success rate as the percentage of successful tests out of the total.
To distinguish the characteristics of \textit{pattern-based} and \textit{categorical reasoning}, we evaluate these results separately.


\begin{table}[h]
\caption{Ablation study comparing DynaCon's prompt with unstructured prompt.}
\vspace{-10pt}
\begin{center}
\begin{tabular}{|c|ccc|cc|}
\hline
\text{Scenario}                                                             & \multicolumn{3}{c|}{1 (pattern)}                                    & \multicolumn{2}{c|}{2 (category)}                           \\ \hline
\text{time-step}                                                                 & \multicolumn{1}{c|}{1}   & \multicolumn{1}{c|}{2}   & 3   & \multicolumn{1}{c|}{1}            & 2            \\ \hline
\text{\begin{tabular}[c]{@{}c@{}}Unstructured \\ Prompt\end{tabular}}       & \multicolumn{1}{c|}{-}   & \multicolumn{1}{c|}{-}   & 202 & \multicolumn{1}{c|}{-}            & -            \\ \hline
\text{\begin{tabular}[c]{@{}c@{}}DynaCon's \\ Prompt\\ (Ours)\end{tabular}} & \multicolumn{1}{c|}{\textbf{205}} & \multicolumn{1}{c|}{\textbf{203}} & \textbf{202} & \multicolumn{1}{c|}{\textbf{refrigerator}} & \textbf{refrigerator} \\ \hline
\end{tabular}
\label{tab1}
\end{center}
\vspace{-10pt}
\end{table}

\section{Results}
\label{sec:results}

\subsection{Ablation Study for Prompt Engineering}
Table~\ref{tab1} summarizes the results of a comparison between two prompts.
In contrast to the prompt designed in DynaCon, the unstructured one failed to capture the dynamic property from the object server.
Even when both prompts were given with the third statement that \textsl{The object list is provided whenever it undergoes changes}, the unstructured prompt did not track the mobile robot's trajectory in real time.
Specifically, in the first scenario, which instructs the robot to find the room number, the unstructured prompt solely focused on the desired goal (Room 202), and did not provide any information about the robot's required actions at each time step.
In the second scenario, it became evident that the  unstructured prompt could not even locate the object or its associated room.
Therefore, our well-structured prompt outperforms the unstructured prompt and is better suited for DynaCon to adjust the navigation path with real-time feedback.

\begin{figure}[t]
\includegraphics[width=\linewidth]{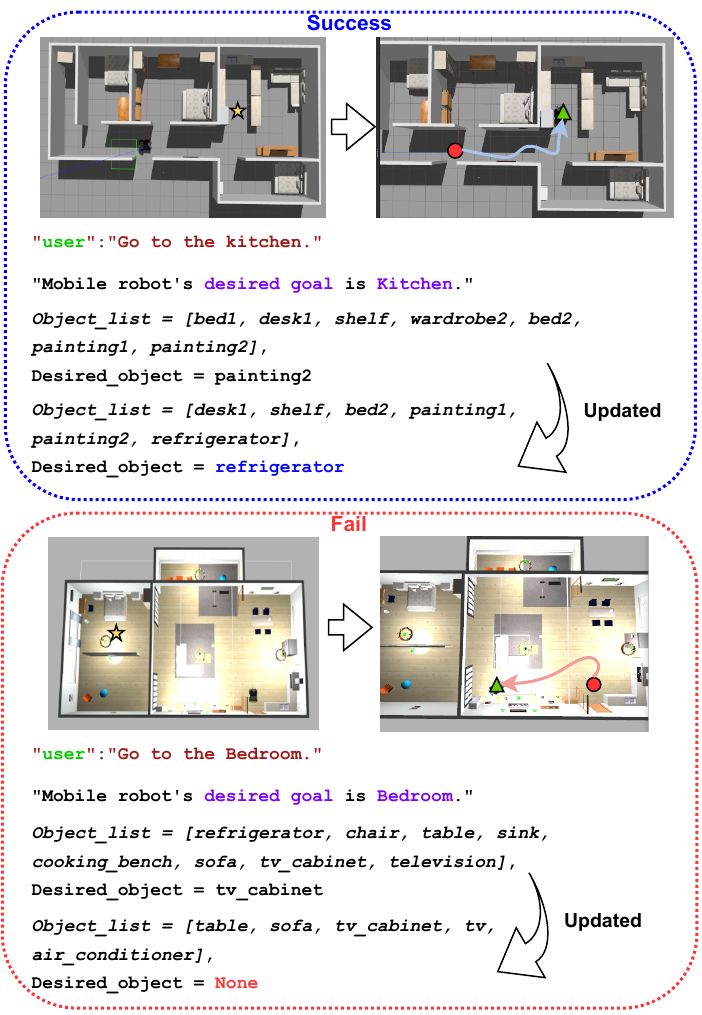}
\caption{Results of success and failure in a virtual house environment. Stars represents the goals, circles denote the initial stages, and triangles indicate the positions at the end of navigation. The user instructed the main task to navigate to specific rooms, and DynaCon provided real-time updates on the object list and desired object. In the failed case, it gave \textit{None} as the desired object, signifying that the goal was not reached.}
\vspace{-17pt}
\label{fig6}
\end{figure}

\subsection{Experiments}
Table~\ref{tab2} represents the results of simulated experiments at the corridor to evaluate the ability of \textit{pattern-based reasoning}.
The mobile manipulator at every test succeed to find the path to the desired numbers.
The positions T1, T3 at Test 1 and T4, T6 at Test 2 start at the beginning of a series of numbers which is simply increased or decreased.
And the positions T2 at Test 1 and T5 at Test 2 are located at the center of the series.
It can be confirmed that regardless of the position in the pattern in which the robot is located, there is no difficulty in performing context-aware navigation.
Thus, in the case of common patterns which we can see in real life, DynaCon is able to estimate the position of goal labeled by digits even though the numbers are not recognized yet by the sensor.

\begin{table}[h]
\caption{Experiment results for \textit{pattern-based reasoning}.}
\vspace{-10pt}
\begin{center}
\begin{tabular}{|c|c|c||c|c|}
\hline
\cline{1-5} 
\hline
\text{Test} & \text{World}& \text{Initial Point}& \text{Task$^{*}$} & \text{Success Rate(\%)} \\
\hline
1 & Corridor 1 & T1 & \cmark & \multirow{8}{*}{100} \\
2 & Corridor 1 & T2 & \cmark &   \\
3 & Corridor 1 & T3 & \cmark &   \\
4 & Corridor 2 & T4 & \cmark &   \\
5 & Corridor 2 & T5 & \cmark &  \\
6 & Corridor 2 & T6 & \cmark &  \\
7 & Corridor 3 & T7 & \cmark &  \\
8 & Corridor 3 & T8 & \cmark &  \\
\hline
\multicolumn{5}{r}{$^{*}$(\cmark : success / \xmark : fail)}
\end{tabular}
\label{tab2}
\end{center}
\vspace{-10pt}
\end{table}

Table~\ref{tab3} presents the results indicating whether the robot successfully located the appropriate room based on desired object.
In some instances, the mobile robot manipulator failed to reach its destination, resulting in an overall success rate of 62.5\% for navigation task.
The failed situations involved different goals, including the home gym in House 1 (Test 3), the bedroom in House 1 (Test 4), and the living room in House 3 (Test 8).
Notably, tests conducted at House 2 had no erroneous cases, as House 2 did not have partitions dividing the rooms, making it easier for the sensor to detect objects related to the target.

The failed cases were primarily attributed to the initial object list, which did not contain items relevant to to the input destination.
For instance, in Test 4, DynaCon received the main task of reaching the bedroom, but the initial object list was represented as \{refrigerator, chair, table, sink, cooking\textunderscore bench, sofa, tv\textunderscore cabinet, television\}.
None of these objects were related to the bedroom, leading DynaCon to select a desired object that it deemed similar.
As the mobile robot continued to move, the desired object remained mismatched with the destination room, ultimately causing the robot to navigate to the wrong target.

However, it is worth noting that the real-time feedback provided by DynaCon could correct the path in the right direction.
In Test 7, the initial object list \{bed, desk, shelf, wardrobe, painting\} did not include kitchen-related objects. After some time of  mobile robot movement, the list was updated to include the refrigerator, enabling the robot to navigate toward the actual location in the kitchen. The process and driving paths for both successful and failed attempts to reach the goal are depicted in Fig.~\ref{fig6}.

\begin{table}[h]
\caption{Experiment results for \textit{categorical reasoning}.}
\vspace{-10pt}
\begin{center}
\begin{tabular}{|c|c|c||c|c|}
\hline
\cline{1-5} 
\hline
\text{Test} & \text{World}& \text{Goal}& \text{Task$^{*}$} & \text{Success Rate(\%)} \\
\hline
1 & House 1 & Kitchen & \cmark & \multirow{8}{*}{62.5} \\
2 & House 1 & Living Room & \cmark &   \\
3 & House 1 & Home Gym & \xmark &   \\
4 & House 1 & Bedroom & \xmark &   \\
5 & House 2 & Kitchen & \cmark &  \\
6 & House 2 & Living Room & \cmark &  \\
7 & House 3 & Kitchen & \cmark &  \\
8 & House 3 & Living Room & \xmark &  \\
\hline
\multicolumn{5}{r}{$^{*}$(\cmark : success / \xmark : fail)}
\end{tabular}
\label{tab3}
\end{center}
\vspace{-17pt}
\end{table}

Depending on the complexity of the reasoning task, we conducted  virtual experiments in the order of \textit{pattern-based reasoning} to \textit{categorical reasoning}.
DynaCon demonstrated a high success rate in \textit{pattern-based reasoning}.
In the simulated corridors, DynaCon successfully reached its destination by reasoning the patterns between each object, even for objects that were currently undetected.
However, DynaCon exhibited relatively lower performance in categorizing household objects.
This is because all numeric objects for pattern recognition shared the same rules, but classification required higher-order reasoning due to the objects that did not follow identical rules.
To increase the success rate in diverse and complex environments, it is advised to enhance the ability of \textbf{Indirect Reasoning} to establish relationships between objects that do not share the same rules or between different spaces.


\section{Conclusion}
\label{sec:conclusion}
In this paper, we introduced DynaCon, an integrated system that harnesses LLM to encompass object feedback, contextual awareness, and navigation via ROS. Initially, the object server updates an object list based on the objects detected from robot's current position. Using this object list, we designed a prompt that is divided into role, main task, and instructions with structured engineering skills for reasoning the environment and its goal.
We also made each portion as a module so that the customized algorithm for navigation can be used instead of default \textit{move\textunderscore base}.
We conducted the experiments using DynaCon in 8 scenarios for \textit{pattern-based}, and\textit{categorical reasoning} each.
Our experiments showed the possibility of applying LLM for navigation under unknown environments which requires contextual awareness for path planning.

To improve and consolidate DynaCon's ability, we need further research.
First, for robustness, more detailed structure at instructions can be added by prompt engineering.
Since most of the information about surroundings is presented at this part, enhanced design of instruction will reinforce the reasoning, especially on categorization.
Second, we can think about an exceptional case when there is no object at all and object list starts from the empty list.
We can use the \textit{frontier\textunderscore exploration} to solve exceptional case when object list is empty.
But we believe that it would be better to consider the solution using LLM's contextual reasoning for more efficient navigation.
And third, most importantly, to ensure the \textit{categorical reasoning} of DynaCon, we have to solve the problem of \textbf{indirect reasoning} requiring inference from nearby objects which are not directly related to desired category.
DynaCon is a first stage to accomplish the contextual navigation under unknown environment via LLM.
Future works with object recognition as well as previous improvements will be followed to push DynaCon to the next level.

\bibliographystyle{IEEEtran}
\bibliography{references}
\end{document}